\documentclass[runningheads]{llncs}
\usepackage{graphicx}
\usepackage{amsmath,amssymb} 
\usepackage{color}
\usepackage{multirow}

\begin{document}
\title{Temporal Saliency Adaptation\\in Egocentric Videos}
%
%
\author{
Panagiotis Linardos\inst{1}
\and
Eva Mohedano\inst{2}
\and
Monica Cherto\inst{2}
\and
\\
Cathal Gurrin\inst{2}
\and
Xavier Giro-i-Nieto\inst{1}}

\authorrunning{P. Linardos et al.}
%
\institute{Universitat Politecnica de Catalunya, 08034 Barcelona, Catalonia/Spain \and
Dublin City University, Glasnevin, Whitehall, Dublin 9, Ireland \\
\email{linardos.akis@gmail.com, xavier.giro@upc.edu}}
\maketitle              
\begin{abstract}
This work adapts a deep neural model for image saliency prediction to the temporal domain of egocentric video. 
We compute the saliency map for each video frame, firstly with an off-the-shelf model trained from static images, secondly by adding a a convolutional or conv-LSTM layers trained with a dataset for video saliency prediction.
We study each configuration on EgoMon, a new dataset made of seven egocentric videos recorded by three subjects in both free-viewing and task-driven set ups.
Our results indicate that the temporal adaptation is beneficial when the viewer is not moving and observing the scene from a narrow field of view.
Encouraged by this observation, we compute and publish the saliency maps for the EPIC Kitchens dataset, in which viewers are cooking. 
\end{abstract}
\section{Motivation}

Saliency prediction refers to the task of estimating which regions of an image have a higher probability of being observed by a viewer.
The result of such predictions is expressed under the form of a saliency map (heat maps), in which higher values are aligned with those pixel locations with higher probabilities of attracting the viewer's attention.
This information can be used for multiple applications, such as a higher quality coding of the salient regions \cite{zhu2018spatiotemporal}, spatial-aware feature weighting \cite{reyes2016my}, or image retargeting \cite{theis2018faster}.
This task has been extensively explored in set ups where the viewer is asked to observe an image \cite{Itti1998PAMI,harel2006nips,judd2009iccv,borji2012cvpr} or video \cite{Wang2018a} depicting a scene.


Our work focuses on the case of egocentric vision, which presents the particularity of having the viewer immersed in the scene.
In this case, the user is not only free to fixate the gaze over any region, but also to change the framing of the scene with his head motion.
When collecting datasets, this set up also differs from others in which the same image or video is shown to many viewers, as in this case each recording and scene is unique for each user.
Egocentric saliency prediction has been studied in the past \cite{Fathi2012,UTego}, a research line that we extend by assessing a state of the art model in image saliency prediction to this egocentric video set up.
We developed our study on a new egocentric video dataset, named \textit{EgoMon}, and added a temporal adaptation on the SalGAN model~\cite{Pan2017} for image saliency prediction. 
We observe that the temporal saliency adaptation improves performance when the viewer is engaged in a task and with a narrow field of view, but, on the other hand, losses are measured when the viewer is simply free-viewing an open scene. 
Encouraged by these results, we have computed the saliency maps pertaining to the Epic Kitchens object detection challenge \cite{EPICKITCHENS}. We believe that these data can be valuable for third-party research focusing on other task such as object detection \cite{reyes2016my} or video summarization \cite{xu2015gaze}. 
Both the EgoMon dataset, Epic Kitchens saliency maps and trained models are publicly available \footnote{\url{https://imatge-upc.github.io/saliency-2018-videosalgan/}}.

\section{The EgoMon Gaze and Video Dataset}




The recording of an egocentric video dataset requires a wearable camera, but also a wearable eye tracker. This specificity in the hardware, together with the privacy constraints, limits the availability of public datasets in this domain.
The GTEA Gaze dataset was collected using Tobii eye-tracker glasses \cite{Fathi2012}. The more updated version of the dataset (EGTEA+) contains 28 hours of cooking activities from 86 unique sessions of 32 subjects. 
Similarly, the University of Texas at Austin Egocentric (UT Ego) Dataset \cite{UTego} was collected using the Looxcie wearable (head-mounted) camera. It contains four videos, each video 3-5 hours long and captured in a natural, uncontrolled setting. The videos depict a variety of activities such as eating, shopping, attending a lecture, driving, or cooking. 




In this work we introduce \textit{EgoMon}, a new egocentric gaze and video dataset.
Data was recorded in Dublin (Ireland) by three different individuals wearing a pair of Tobii glasses equipped with a monocular eye tracker. The dataset is delivered as a collection of seven videos of an average length of 30 minutes. 
EgoMon includes both \textit{free-viewing activities} (a walk in a park, walking to the office, a walk in the botanic gardens, a bus ride), as well as \textit{task-oriented activities} (cooking an omellette, listening to an oral presentation and playing cards). In the case of the botanic gardens, an additional 
a sequence of images captured every 30 seconds with a Narrative clip camera is also provided.


\section{Deep Neural Models for Temporal Saliency Adaptation}

Video saliency prediction with deep neural networks has basically adapted to this task the architectures proposed for video action recognition.
Two-stream networks \cite{simonyan2014two} combining video frames and optical flow were applied in \cite{bak2017spatio} for saliency prediction, while temporal sequences modeled with RNN \cite{donahue2015long} were adopted in \cite{jiang2017predicting}. 
The authors of the largest dataset for video saliency prediction, the DHF1K (Dynamic Human Fixation 1K) dataset\cite{Wang2018a}, also trained a deep neural model based on ConvLSTM layers to predict the saliency maps.
Similarly, the authors of \cite{gorji2018going} propose a complex convolutional architecture with four branches fused with a temporal-aware ConvLSTM layer. 
Regarding egocentric saliency prediction with deep models, Huang \emph{et al.}~\cite{huang2018predicting} propose to model the bottom-up and top-down attention mechanisms on the GTEA Gaze dataset. Their approach combines a saliency prediction with a task-dependent attention module, which explicitly models the temporal shift of gaze fixations during different manipulation tasks.

Our proposed architecture starts by processing each video frame separately with SalGAN \cite{Pan2017}, an image-based saliency prediction pre-trained trained on the SALICON dataset \cite{huang2015salicon}.
SalGAN outputs a sequence of static saliency maps which were fed into two types of adaptation layers: 128 convolutional filters \cite{lecun1998gradient} of kernel size 3x3 and padding of 1, and its temporal-aware counterpart as ConvLSTM \cite{ConvLSTM} with the same convolutional parameters. Their parameters were estimated from 700 training videos from the DHF1K dataset \cite{Wang2018a}. 
An SGD optimizer with 0.9 momentum was used, and the learning rate started at 0.00001 and decayed with a 0.1. There was also a weight decay of 0.0001.

\begin{figure}
\includegraphics[width=\textwidth]{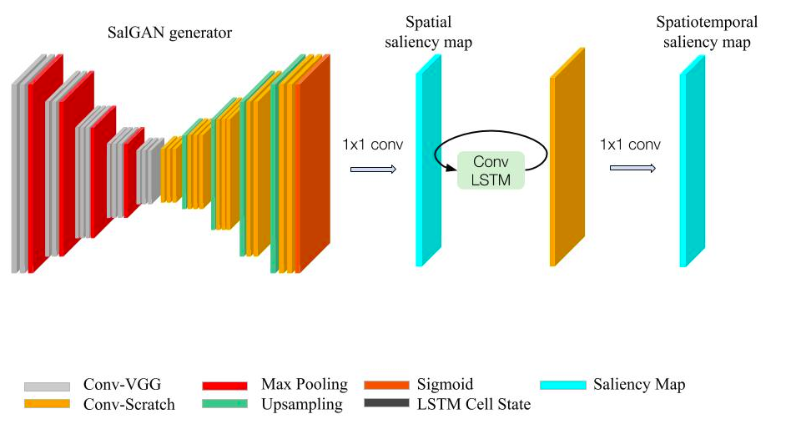}
\caption{Architecture of the dynamic model. The static model uses plain convolutions without the LSTM temporal recurrence.} 
\label{fig2}
\end{figure}

\section{Experimentation}

The proposed model was assessed firstly on the same DHF1K dataset \cite{Wang2018a} the same from which the conv and convLSTM layers were trained. 
Afterwards, the model was assessed on the proposed EgoMon dataset to draw our conclusions in the egocentric domain.

\begin{table}[t]
\centering
\caption{Performance on the DHF1K dataset.}
\label{tab:dhf1k}
\begin{tabular}{|c|c|c|c|c|c|}
\hline
 &\hspace{0.1cm} AUC-J $\uparrow$ \hspace{0.1cm} &\hspace{0.1cm}sAUC $\uparrow$\hspace{0.1cm} & \hspace{0.1cm}NSS $\uparrow$\hspace{0.1cm} 	& \hspace{0.1cm} CC $\uparrow$\hspace{0.1cm} & \hspace{0.1cm}SIM $\uparrow$\hspace{0.1cm}\\
\hline
SoA \cite{Wang2018a} & 0.885 & 0.553	&  2.259 &	\textbf{0.415} & \textbf{0.311}\\

SalGAN ~\cite{Pan2017} & \textbf{0.930} & \textbf{0.834} &	\textbf{2.468} & 0.372 & 0.264  \\
+ conv & 0.743	& 0.723	&  2.208 &	0.303 & 0.261\\
+convLSTM & 0.744	& 0.722	&  2.246 &	0.302 & 0.260\\
\hline
\end{tabular}
\end{table}

Table \ref{tab:dhf1k} indicates that, surprisingly, the off-the-shelf (frame-based) SalGAN model ~\cite{Pan2017} outperformed the state of the art model on the DHF1K \cite{Wang2018a} dataset. 
On the other hand, the quality of the prediction decreases when the conv or convLSTM layers are trained on top, which indicates that the domain adaptation is damaging the performance of the original SalGAN. 

\begin{table}[t]
\centering
\caption{NSS metric across the DHF1K and EgoMon datasets.}
\label{tab:egomon}
\begin{tabular}{|c|c|c|c|}
\hline
 &\hspace{0.1cm}  DHF1K \hspace{0.1cm} & \hspace{0.1cm}EgoMon\hspace{0.1cm} \\
\hline
SalGAN ~\cite{Pan2017} & \textbf{2.468}  & \textbf{2.079}\\
+conv  & 2.208  & 1.250\\
+convLSTM &  2.246  & 1.247\\
\hline
\end{tabular}
\end{table}

\begin{table}
\centering
\caption{Performance on different EgoMon tasks (NSS metric). }\label{tab3}
\label{tab:FreeviewVsTasks}
\begin{center}
\begin{tabular}{ |c|c|c|c|c|c| } 

\hline
\multirow{2}{4em}{} & \multicolumn{5}{|c|}{free-viewing recordings (bottom-up saliency)}  \\
\cline{2-6}
& bus ride& botanical gardens & dcu park & walking office & AVERAGE\\ 
\hline
SalGAN ~\cite{Pan2017} & \textbf{1.618} & \textbf{1.182} & \textbf{4.374} &	\textbf{3.435} & \textbf{2.652} \\
+ conv & 0.947 & 0.846 & 0.683 & 0.745 & 0.805 \\
+ convLSTM & 0.827 & 0.576 & 1.172 & 1.040 & 0.904 \\
\hline
\multirow{2}{4em}{} & \multicolumn{5}{|c|}{task-driven recordings (top-down saliency)}  \\
\cline{2-6}
& playing cards & presentation & tortilla & &AVERAGE\\
\hline
SalGAN ~\cite{Pan2017} & 0.967 & 1.360 & 1.618 && 1.315\\
+ conv 					& 1.114 & \textbf{1.966} & 2.002 && 1.694\\
+ convLSTM 				& \textbf{1.141} & 1.897 & \textbf{2.077} && \textbf{1.705}\\
\hline
\end{tabular}
\end{center}
\end{table}

Table \ref{tab:egomon} indicates an even worse loss of performance when adding this adaptation layers in the EgoMon dataset.
Nevertheless, the more detailed look provided in Table \ref{tab:FreeviewVsTasks}
that actually the adaptation layers are beneficial in those scenes where the user is engaged in an activity.

Qualitative analysis of the saliency maps showed that the convolutional layers (with and without temporal information) had the effect of reinforcing the higher probability pixels at the expense of darkening the lower ones. This effect beneficial in the case of task-driven activities, because the scene tends to be constant in time and the region of interest is localized in the space. However, free-viewing tasks contain changing scenes with much more sparse saliency maps. 

\section{Ackowledgements}

Panagiotis Linardos and Monica Cherto were supported by the Erasmus+ Program from the European Union for student mobility.
This research was partially supported by the Spanish Ministry of Economy and Competitivity and the European Regional Development Fund (ERDF) under contract TEC2016-75976-R. 
We acknowledge the support of NVIDIA Corporation for the donation of GPUs.

%
%
%
\bibliographystyle{splncs04}
\bibliography{biblio.bib}

\begin{thebibliography}{10}
\providecommand{\url}[1]{\texttt{#1}}
\providecommand{\urlprefix}{URL }
\providecommand{\doi}[1]{https://doi.org/#1}

\bibitem{bak2017spatio}
Bak, C., Kocak, A., Erdem, E., Erdem, A.: Spatio-temporal saliency networks for
  dynamic saliency prediction. IEEE Transactions on Multimedia  (2017)

\bibitem{borji2012cvpr}
Borji, A.: Boosting bottom-up and top-down visual features for saliency
  estimation. In: IEEE Conference on Computer Vision and Pattern Recognition
  (CVPR) (2012)

\bibitem{EPICKITCHENS}
Damen, D., Doughty, H., Farinella, G.M., Fidler, S., Furnari, A., Kazakos, E.,
  Moltisanti, D., Munro, J., Perrett, T., Price, W., Wray, M.: Scaling
  egocentric vision: The epic-kitchens dataset. In: European Conference on
  Computer Vision (ECCV) (2018)

\bibitem{donahue2015long}
Donahue, J., Anne~Hendricks, L., Guadarrama, S., Rohrbach, M., Venugopalan, S.,
  Saenko, K., Darrell, T.: Long-term recurrent convolutional networks for
  visual recognition and description. In: Proceedings of the IEEE conference on
  computer vision and pattern recognition. pp. 2625--2634 (2015)

\bibitem{Fathi2012}
Fathi, A., Li, Y., Rehg, J.M.: {Learning to recognize daily actions using
  gaze}. Lecture Notes in Computer Science (including subseries Lecture Notes
  in Artificial Intelligence and Lecture Notes in Bioinformatics)  \textbf{7572
  LNCS}(PART 1),  314--327 (2012)

\bibitem{gorji2018going}
Gorji, S., Clark, J.J.: Going from image to video saliency: Augmenting image
  salience with dynamic attentional push. In: Proceedings of the IEEE
  Conference on Computer Vision and Pattern Recognition. pp. 7501--7511 (2018)

\bibitem{harel2006nips}
Harel, J., Koch, C., Perona, P.: Graph-based visual saliency. In: Neural
  Information Processing Systems (NIPS) (2006)

\bibitem{huang2015salicon}
Huang, X., Shen, C., Boix, X., Zhao, Q.: Salicon: Reducing the semantic gap in
  saliency prediction by adapting deep neural networks. In: IEEE International
  Conference on Computer Vision (ICCV) (2015)

\bibitem{huang2018predicting}
Huang, Y., Cai, M., Li, Z., Sato, Y.: Predicting gaze in egocentric video by
  learning task-dependent attention transition. arXiv preprint arXiv:1803.09125
   (2018)

\bibitem{Itti1998PAMI}
Itti, L., Koch, C., Niebur, E.: A model of saliency-based visual attention for
  rapid scene analysis. IEEE Transactions on Pattern Analysis and Machine
  Intelligence (PAMI) (20),  1254–--1259 (1998)

\bibitem{jiang2017predicting}
Jiang, L., Xu, M., Wang, Z.: Predicting video saliency with object-to-motion
  cnn and two-layer convolutional lstm. arXiv preprint arXiv:1709.06316  (2017)

\bibitem{judd2009iccv}
Judd, T., Ehinger, K., Durand, F., Torralba, A.: Learning to predict where
  humans look. In: IEEE International Conference on Computer Vision (ICCV)
  (2009)

\bibitem{lecun1998gradient}
LeCun, Y., Bottou, L., Bengio, Y., Haffner, P.: Gradient-based learning applied
  to document recognition. Proceedings of the IEEE  \textbf{86}(11),
  2278--2324 (1998)

\bibitem{Pan2017}
Pan, J., Ferrer, C.C., McGuinness, K., O'Connor, N.E., Torres, J., Sayrol, E.,
  Giro-i Nieto, X.: {SalGAN: Visual Saliency Prediction with Generative
  Adversarial Networks}  (2017)

\bibitem{reyes2016my}
Reyes, C., Mohedano, E., McGuinness, K., O'Connor, N.E., Giro-i Nieto, X.:
  Where is my phone?: Personal object retrieval from egocentric images. In:
  Proceedings of the first Workshop on Lifelogging Tools and Applications. pp.
  55--62. ACM (2016)

\bibitem{ConvLSTM}
Shi, X., Chen, Z., Wang, H., Yeung, D., Wong, W., Woo, W.: Convolutional {LSTM}
  network: {A} machine learning approach for precipitation nowcasting. CoRR
  \textbf{abs/1506.04214} (2015), \url{http://arxiv.org/abs/1506.04214}

\bibitem{simonyan2014two}
Simonyan, K., Zisserman, A.: Two-stream convolutional networks for action
  recognition in videos. In: Advances in neural information processing systems.
  pp. 568--576 (2014)

\bibitem{UTego}
Su, Y.C., Grauman, K.: Detecting engagement in egocentric video. In: European
  Conference on Computer Vision. pp. 454--471. Springer (2016)

\bibitem{theis2018faster}
Theis, L., Korshunova, I., Tejani, A., Husz{\'a}r, F.: Faster gaze prediction
  with dense networks and fisher pruning. arXiv preprint arXiv:1801.05787
  (2018)

\bibitem{Wang2018a}
Wang, W., Shen, J., Guo, F., Cheng, M.M., Borji, A.: Revisiting video saliency:
  A large-scale benchmark and a new model. In: Proceedings of the IEEE
  Conference on Computer Vision and Pattern Recognition. pp. 4894--4903 (2018)

\bibitem{xu2015gaze}
Xu, J., Mukherjee, L., Li, Y., Warner, J., Rehg, J.M., Singh, V.: Gaze-enabled
  egocentric video summarization via constrained submodular maximization. In:
  Proceedings of the IEEE Conference on Computer Vision and Pattern
  Recognition. pp. 2235--2244 (2015)

\bibitem{zhu2018spatiotemporal}
Zhu, S., Xu, Z.: Spatiotemporal visual saliency guided perceptual high
  efficiency video coding with neural network. Neurocomputing  \textbf{275},
  511--522 (2018)

\end{thebibliography}
\end{document}